\documentclass{article} 
\usepackage{iclr2015,times}
\usepackage{hyperref}
\usepackage{url}
\usepackage[pdftex]{graphicx}
	\graphicspath{{./image/}}
   	\DeclareGraphicsExtensions{.pdf,.jpeg,.png}
   	
 \usepackage{amsmath}
 
\title{Bach in 2014: Music Composition with Recurrent Neural Network}

\author{
I-Ting Liu 
\\
Department of Music\\
Carnegie Mellon University\\
Pittsburgh, PA 15213, USA \\
\texttt{itingl@andrew.cmu.edu} \\
\And
Bhiksha Ramakrishnan \\
Language Technologies Institute \\
Carnegie Mellon University \\
Pittsburgh, PA 15213, USA \\
\texttt{bhiksha@cs.cmu.edu} \\
}

\begin{document}

\maketitle

\begin{abstract}
We propose a framework for computer music composition that uses resilient propagation (RProp) and long short term memory (LSTM) recurrent neural network. In this paper, we show that LSTM network learns the structure and characteristics of music pieces properly by demonstrating its ability to recreate music. We also show that predicting existing music using RProp outperforms Back propagation through time (BPTT). 
    
\end{abstract}

\section{Introduction}
Composing music with computer has attracted researchers for a long time. Among different approaches, artificial neural networks have been proposed to handle this task, as neural networks were originally developed to model how human brain works. Neural networks have been very successful in language modeling, pattern recognition, and predicting time series data. Music generation can be formulated as a time-series prediction problem: the note played at each time can be regarded as a prediction given the notes played before. 

The most common neural network is a multi-layered feed-forward network. The network predicts the next note played in time given the previous note played. This kind of neural network has a limited ability to capture rhythmic pattern and music structure in music as it does not have a mechanism to keep track of the notes played in the past. On the other hand, the fact that music is a complex structure that has both short-term and long-term dependency just as a language model makes recurrent neural network (RNN) an ideal structure for this task. Feedback connections in RNN enable RNN to maintain an internal state for temporal relationship of the inputs. 

However, RNN has been notoriously hard to train because of "vanishing gradients, \citep{hochreiter1997long}" a problem commonly seen in RNNs when training with gradient based methods. Gradient methods, such as Back-Propagation Through Time (BPTT) \citep{werbos1990backpropagation}, Real-Time Recurrent Learning (RTRL) \citep{robinson1987utility} and their combination, update the network by flowing errors "back in time." As the error propagates from layer to layer, it tends to either explode or shrink exponentially depending on the magnitude of the weights. Therefore, the network fails to learn learn long-term dependency between inputs and outputs. Tasks with time lags that are greater than 5-10 are already difficult to learn \citep{hochreiter1997long}, not to mention that dependency of music usually spans across tens to hundreds of notes in time, which contributes to music's unique phrase structures. 

Long short term memory (LSTM) \citep{hochreiter1997long} algorithm was designed to tackle the error-flow problem by enforcing constant error flow through "constant error carousels" in internal states. LSTM learns quickly and efficiently, and is proved to be effective in multiple recognition tasks. \citet{eck2002first} was the first that employed LSTM and RNN to learn and compose blues music. \citet{franklin2006recurrent} later on utilized LSTM networks to model Jazz music. We believe LSTM recurrent neural networks could learn the global structure of music pieces well, but the efficiency and efficacy of the training phase could be further improved with adaptive learning algorithm. 

In this paper, we propose a framework for music composition that uses resilient propagation (RPROP) \citep{riedmiller1993direct} \citep{Riedmiller94rprop} in replace of standard back propagation to train the recurrent neural network. We will also use long short term memory cells in the network as they have better ability to learn songs, long song phrases and structrues precisely.

The remainder of the paper is organized as follows. In section~\ref{sec: relatedWork} we will look at past works that use RNN for computer-aided music generation. In section~\ref{sec:method}, we will introduce our system framework. We will then evaluate the system by conducting experiments in section~\ref{sec:experiment}.

\section{Related Work}
\label{sec: relatedWork}
\citet{todd} was one of the earliest paper that used RNN for note-by-note music generation with Jordan recurrent neural network. Jordan recurrent network is a simple RNN that has a recurrent connection from the output layer to the input layer, and a self-recurrent link at the input layer.   The network is trained with BPTT, and recurrence is managed by teacher-forcing. In the training phase, Todd trained monophonic melodies with the network. The trained network could then be used to generate music by either mixing and varying the original training data, or by introducing new "seed melody" as the input, and the rest of the network output are recorded as the generated music. 

In \citet{mozer}, a fully connected RNN was trained by minimizing log-likelihood function of the L2 norm of the predicted and actual output via back-propagation through time (BPTT). The outputs of the final layer are treated as probability of whether the note should be on or off. In addition, to better model harmonic relationship of musical notes, Mozer proposed a grey-code like representation that encodes notes based on their location on chromatic circle, circle of fifths and pitch height, a psychologically based representation derived from \citet{shepard1982geometrical}. To compensate BPTT-trained RNN's inefficacy of learning long-term dependencies, Mozer also used a similar encoding scheme to represent durations based on three fraction scales. 

\citet{franklin} adopted Todd's network and added a second training phase, where the network was further trained via reinforcement learning. In reinforcement learning phase, a scalar value was calculated by a set of "music rules" to determine how good the output is, and is then used to replace the explicit error information. 
 
To deal with vanishing gradient problem, \citet{eck2002first} used two long short-term memory (LSTM) networks, one for learning melody and one for chords, to compose blues music. The output of the chord network is connected to the input of the melody network. The system was able to learn the standard 12-bar blues chord sequences and generate music notes that follows the chords. \citet{franklin2006recurrent} also used LSTM networks to learn Jazz music. They developed a pitch representation scheme based on major and minor thirds, the circles-of-thirds representation, inspired by \citet{mozer}'s circle-of-fifths pitch representation. They also extended Mozer's duration representation by dividing note durations into 96 subdivisions, corresponding to a "tick" in the Musical Instrument Digital Interface (MIDI)\citep{messick1988maximum} standard digital protocol. 

To describe music's correlated pattern among multiple notes, \citet{boulanger} developed an RNN-based model by using restricted Bolzmann machine (RBM) \citep{smolensky1986information} and recurrent temporal RBM(RTRBM) \citep{sutskever2009recurrent}. The model, RNN-RBM, allows freedom in describing the temporal dependencies of the notes, and is believed to be able to model unconstrained polyphonic music in a piano-roll representation without any dimension reduction.

\section{Method}
\label{sec:method}

This section goes over each individual component involved in building the whole system. 

\subsection{Music Representation}
\label{subsec:representation}
Multiple input and output neurons are used to represent different pitches. If a note is played at a given time, then the value of the neuron associated with the particular pitch is 1.0, otherwise 0.0. In our system, we use 88 binary visible units that span the whole range of a piano from A0 to C8 as was done by \citet{boulanger}. The reason why we avoided psychologically distributed encodings or any other dimension reduction techniques but instead represent the data in this simple form is that we believe that a good network should be able to identify harmonically correlated pattern between notes by learning bias. Besides, such representation is flexible in representing both monophonic and polyphonic data. 

To represent music on a RNN, we split time into fractions. The length of the fraction depends on the type of the music we are training. 
Computer music composition problem could then be formulated as a supervised time-series problem. The input to the network at time $t$, $\mathbf{x}(t)$, is a 88-dimension vector representing the note played at time $t$. The target at time $t$, $\mathbf{y}(t)$ is the note played at time $t+1$, i.e. $\mathbf{x}(t+1)$. The network is then trained over multiple iterations, each of which learns and predicts the note of the next fraction.   

\subsection{Recurrent Neural Networks and Long Short Term Memory}
\begin{figure}[t!]
\centering
\includegraphics[width=2.5in]{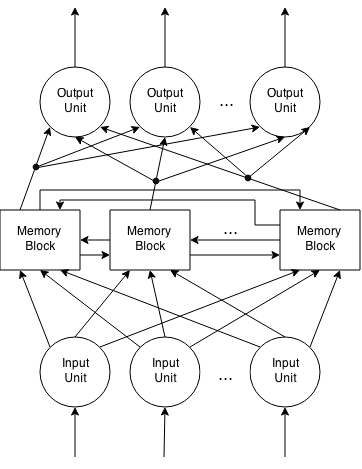}
\caption{Long Short Term Memory}
\label{fig:LSTM}
\end{figure}

A recurrent neural network (RNN) has at least one feedback connection from one or more of its units to units, thus forming cyclic paths in the network. RNNs are known to be able to approximate dynamical systems due to the internal states that act as internal memory to process sequence of inputs through time. In this paper, we use standard long short-term memory structure as shown in Figure~\ref{fig:LSTM}.

The network has one fully-connected hidden layer of memory blocks, which contains one or more units (memory cells). A memory block also contains three sigmoid gating units: input gate, output gate, and forget gate. An input gate learns to control when inputs are allowed to pass into the cell in the memory block so that only relevant contents are remembered; a output gate learns to control when the cell's output should be passed out of the block, protecting other units from interference from current irrelevant memory contents; A forget gate learns to control when it is time forget already remembered value, i.e. to reset the memory cell. The outputs of all memory blocks are fed back recurrently to all memory blocks to "remember" past values. For more detail about LSTM, please refer to \citet{hochreiter1997long}.

\subsection{Resilient Propagation (RProp)}
Training of the long short term memory (LSTM) network is done by resilient propagation (RProp), a heuristic optimization algorithm proposed by \citep{riedmiller1993direct}. It is a kind of local adaptation learning strategy, which aims to facilitate gradient-descent learning problem by modifying weight-specific parameters, such as learning rate, by using only weight-specific information, which is partial derivatives in this case. The basic principle of RProp is to directly adapt the size of the weight update by ignoring the size of the partial derivatives. Unlike other adaptive learning algorithm that takes account of the magnitude of the gradient, only the sign of the derivative is used to perform both learning and adaptation. The direction of the weight update is indicated by the sign of the derivative, and the size of weight change is decided solely by weight-specific update values. The update value $\Delta w_{ij}^{(t)}$ is now given by

$$
\Delta w_{ij}^{(t)}=
\left\{\begin{matrix}
-\Delta _{ij}^{(t)} , &  if  & \frac{\partial E^{(t)}}{\partial w_{ij}} > 0 \\ 
+ \Delta_{ij}^{(t)}, & if &   \frac{\partial E^{(t)}}{\partial w_{ij}} <0\\ 
 0, & else 
\end{matrix}\right.
$$

$$ w_{ij}^{(t+1)} = w_{ij}^{(t)}+ \Delta w_{ij}^{(t)} $$

where $w_{ij}$ is the weight from neuron $j$ to neuron $i$, $E$ is an arbitrary error function, $\frac{\partial E^{(t)}}{\partial w_{ij}}$ is the summed gradient information over all patterns in the pattern set. \citep{riedmiller1993direct}

Researches have shown that networks trained with local adaptive algorithms, especially RProp, converge considerably faster than ordinary gradient descent algorithm \citep{riedmiller1994advanced}. Moreover, RProp does not require parameter tuning in order to obtain optimal results. For more details on implementing RProp, please refer to \citep{Riedmiller94rprop}

\subsection{Learning to compose}
\label{subsec:learningToCompose}
The system is composed of two phases: training phase and testing phase.  In training phase, multiple music pieces are fed into the network using the representation mentioned in section~\ref{subsec:representation}. The network learns multiple music pieces with LSTM recurrent neural network and resilient back-propagation. The goal for both phase is to predict the output probability for each given note to be on. This task is similar to multi-label classification problem: each input instance is assigned with zero or multiple labels, the label being whether that particular note is on or off. The error $E$ is computed with mean square error (MSE), whose definition is given as follows:

$$E = \frac{1}{n}\sum^n_{i=1}(t_i-y_i)^2$$

where $t_i$ is the target value, and $y_i$ is the predicted value. For the output layer, we use a logistic sigmoid layer, which produces output in the range of [0, 1]. 

After the error rate has converged, the network is then tested by starting it with the inputs of the first time step. Outputs for ensuing time steps are then predicted with previous predictions. All neurons whose activation value is greater than a decision threshold, which we empirically set as 0.9, are treated as on, and the notes associated with those neurons are played. The testing phase is also the composition phase, where the predictions could be recorded to form new songs.

\section{Experiments}
\label{sec:experiment}
\subsection{Training Data}
For the experiment in this paper, we used J.S. Bach's Chorale midi dataset and splits by \citet{allan2005harmonising}. The dataset contains 384 four-part harmonization, and was split into training and testing set based on keys (major or minor) of the pieces.
 
\subsection{Music Reconstruction}
For the first part of the experiment, we aim to inspect the network's ability to learn the representation of music pieces. A trained human musician could perform a piece flawlessly after learning and practicing, and we would like to know whether our network is capable of recreating a song as it is after training. 

We randomly picked four chorales from the training subset in the dataset. 
The network was fed one chorale at a time, and was trained until convergence. Then, the beginning notes of each chorale was fed into the network as initial input. The network then predicted notes for ensuing time steps given the previous predicted notes. 

\begin{figure}[h!]
\centering
\includegraphics[width=3.5in]{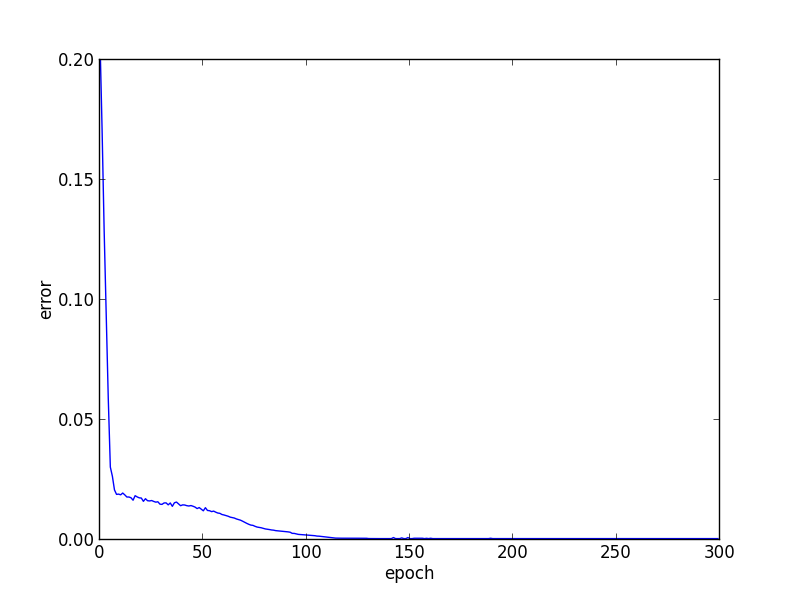}
\caption{Mean square error for one of J.S. Bach's chorales}
\label{fig:34_error}
\end{figure}

\begin{figure}[h!]
\centering
\includegraphics[width=6in]{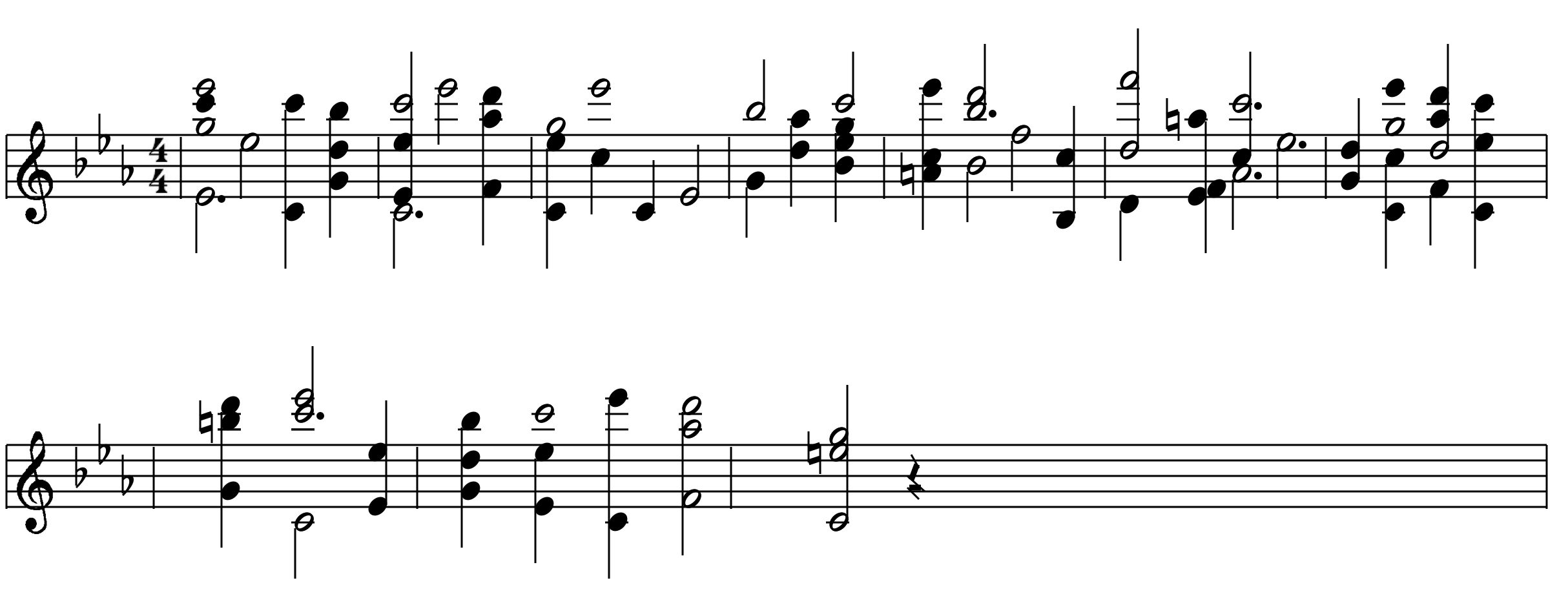}
\caption{Reconstruction of one of J.S. Bach's chorales with LSTM network}
\label{fig:34_pred}
\end{figure}

\begin{figure}[t!]
\centering
\includegraphics[width=6in]{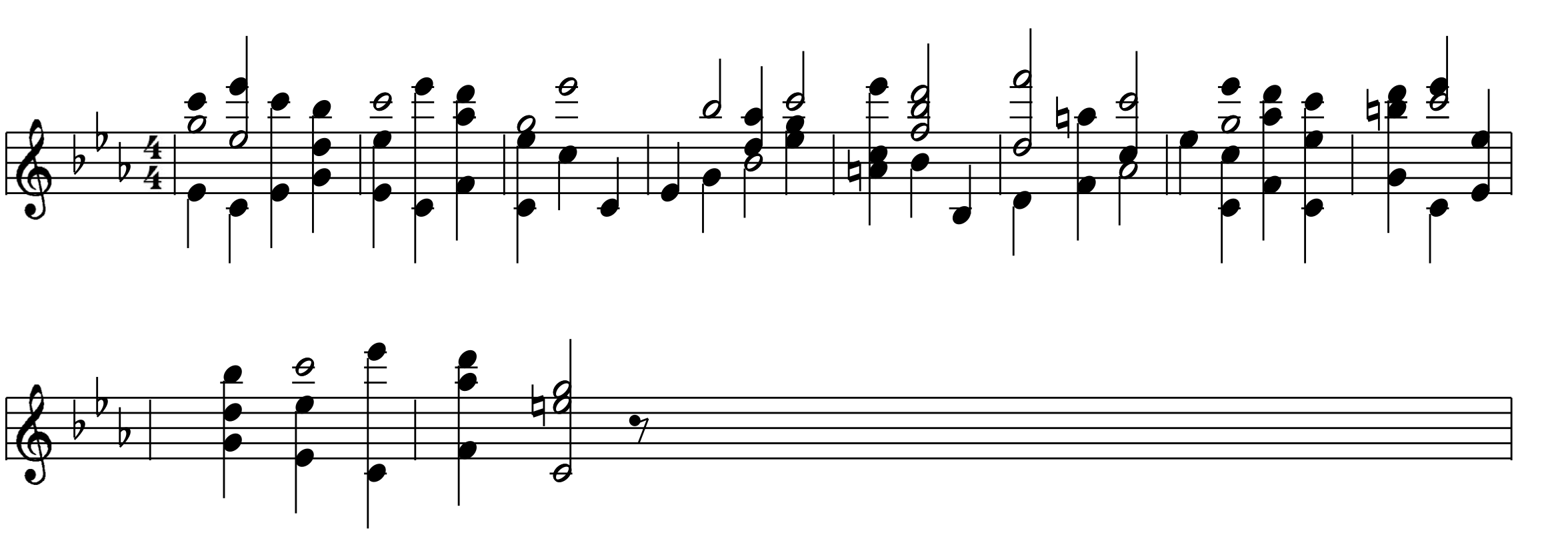}
\caption{Original Score of one of J.S. Bach's chorales}
\label{fig:34_orig}
\end{figure}

The mean square error (MSE) of the training phase for Chorale No.34 is shown as examples in Figure~\ref{fig:34_error}. The network converged exceptionally fast: total MSE reached 1.0\% at epoch 125. Figure~\ref{fig:34_pred} and Figure~\ref{fig:34_orig} show the network's rendition as well as the original score. Note that since we are using midi files as input and output, we did not put emphasis on transcribing midi files nor obtaining  correct time or key signature. The music sheets are merely shown as references of how the network performs. Albeit the reconstruction was not perfect, this experiment demonstrated the network's capability of reconstructing a song fairly well after training. 

\subsection{Music Prediction}

\begin{table}[h]
\centering
\caption{Accuracy and F1 score of J.S. Bach Chorale Dataset with BPTT and RProp}
\label{tab:rprop_vs_bptt}
\begin{tabular}{lll}\\
           & Accuracy    & F1 score     \\
BPTT       & 21.03\% & 11.84\% \\
RProp 	   & 31.91\% & 20.29\%
\end{tabular}
\end{table}

\begin{figure}[h]
\centering
\includegraphics[width=3.5in]{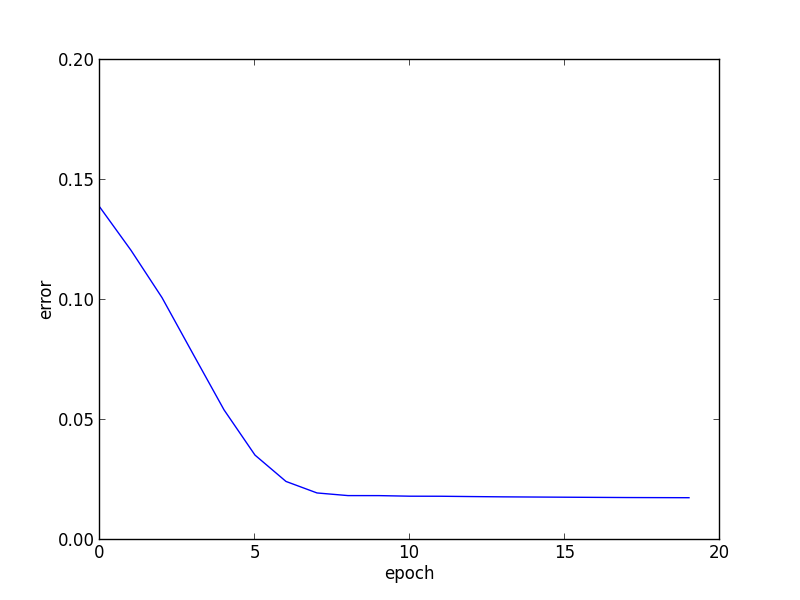}
\caption{Mean square error for one of J.S. Bach's chorales using RProp}
\label{fig:bach_rprop}
\end{figure}

\begin{figure}[h]
\centering
\includegraphics[width=3.5in]{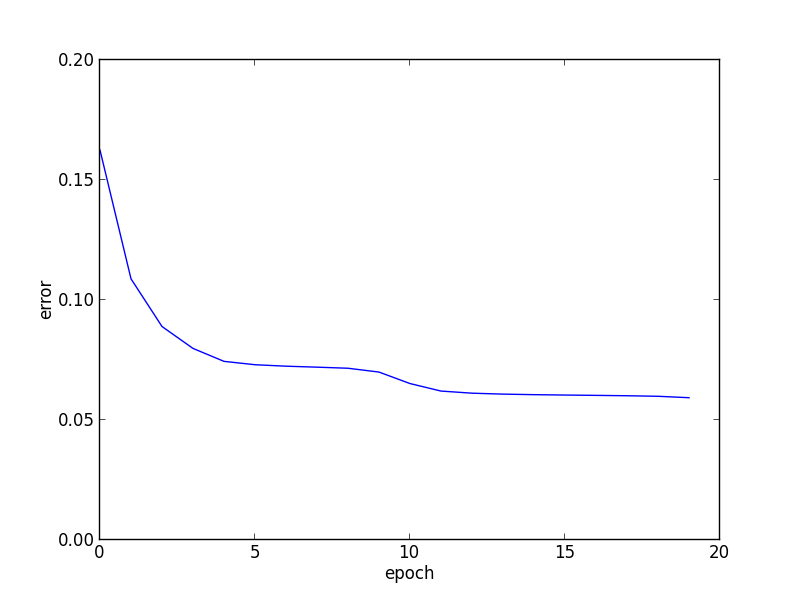}
\caption{Mean square error for one of J.S. Bach's chorales using BPTT}
\label{fig:bach_backprop}
\end{figure}

To see the network's ability to use the knowledge learned during training phase to compose music, we tested the network on the whole J.S. Bach Dataset. As before, MSE of the training phase with RProp is provided in Figure~\ref{fig:bach_rprop}. MSE for the network trained with BPTT is shown in Figure~\ref{fig:bach_backprop}. From the two figures, we could tell that the network also converges rapidly when trained with RProp on the whole dataset compared to the network trained with BPTT. To evaluate performance, we calculated F1 score, a common evaluation scheme for multi label classification task. F1 score $F$, precision $P$ and recall $R$ for each music piece $d_j$ are defined as follows \citep{godbole2004discriminative}:

$$P(d_j) = |T\cap S|/|S|$$ 
$$R(d_j) = |T \cap S|/|T|$$
$$ F1(d_j) = \frac{2P(d_j)R(d_j)}{P(d_j)+R(d_j)} $$ 

We also calculated frame-level accuracy proposed by \citet{bay2009evaluation}. An overall accuracy score $Acc$ is given by

$$
Acc = \frac{\sum^T_{t=1}TP(t)}{\sum^T_{t=1}(TP(t)+FP(t)+FN(t)}
$$

where $TP(t)$, $FP(t)$, $FN(t)$ are true positives, false positives, and false negatives of time (frame index) $t$.

Table~\ref{tab:rprop_vs_bptt} shows the accuracy and f1 score of the network on the testing data when the network was trained with RProp and BPTT. Under the same number of iterations, RProp outperforms BPTT in both metrics.

\section{Discussion}

\subsection{Limitation of Representation}
As mentioned by \citet{eck2002first}, the method we use to represent music ignores two issues. First, it is not possible to separate melody from accompaniment with this kind of representation. While such representation is flexible in representing any kind of data as mentioned earlier in section~\ref{sec:method}, it does not differentiate chords from melody. Second, there is no way to identify when a note ends. Eight eighth notes of the same pitch are represented exactly the same way as four quarter notes of the same pitch. One way to deal with this problem is to shrink the step size and append zero at the end of each note. Another way is to add an additional visible unit in the network to indicate whether the current input is the beginning of a note as done by \citet{todd}. The effectiveness of the two possible implementation should be discussed by conducting more systematic experiments in the future work.  

\subsection{Lack of Proper Evaluation Metric}
In this paper, we evaluated the system with accuracy and F1 score. However, we discovered that these evaluation metrics does not necessarily correspond to human's perception of music similarity. Higher accuracy or F1 score doesn't mean the reconstructed music sounds more similar to the original song. If we are to evaluate his network's reconstruction performance, further study is required to design a better evaluation metric that could capture perceived similarity between music pieces.

\subsection{Further Improvement}
We believe preprocessing the input data could help the network learn better, such as transposing songs to the same key, as suggested by \citet{boulanger}. Also, the system could also be expanded to model  dynamics, which is an important feature of music style. Melody and accompaniment (chords) could also be trained with connected two networks, which might enable us to model music characteristics more comprehensively.

\section{Conclusion}
In this paper, an RNN-based music composition system was proposed. Using LSTM in the network enables the network to learn the structure and rhythm of music pieces, and the information could be used to compose new pieces similar in form. Instead of using BPTT algorithm to train the network, we employed RProp to expedite learning. Experiments showed that the network could learn the music and recreate the original piece well in only tens of iterations (learning epochs). It is also shown in the experiment that the network could compose new music once it learned knowledge about music. 

\subsubsection*{Acknowledgments}
This paper was the project for Deep Learning course offered in Carnegie Mellon University. The author wants to thank course TA, Danny Lan, and instructor, Professor Bhiksha Ramakrishnan, for their great advice.

\bibliography{iclr2015}
\bibliographystyle{iclr2015}

\end{document}